\documentclass[final]{cvpr}

\usepackage{times}
\usepackage{epsfig}
\usepackage{graphicx}
\usepackage{amsmath}
\usepackage{amssymb}

\usepackage{multirow}
\usepackage{algorithm}
\usepackage{algorithmic}
\usepackage{amsfonts}
\usepackage{array}
\usepackage{xcolor,colortbl}
\usepackage{amsmath}
\usepackage{color}

\definecolor{Gray}{gray}{0.85}
\definecolor{LightCyan}{rgb}{0.88,1,1}
\newcolumntype{a}{>{\columncolor{Gray}}c}
\newcolumntype{b}{>{\columncolor{white}}c}

\usepackage[pagebackref=true,breaklinks=true,colorlinks,bookmarks=false]{hyperref}

\def\cvprPaperID{3353} 
\def\confYear{CVPR 2021}

\begin{document}

\title{When Human Pose Estimation Meets Robustness:\\ Adversarial Algorithms and Benchmarks}

\author{
	Jiahang Wang $^{1\dagger}$ \quad  Sheng Jin $^{2,3}$ \quad Wentao Liu$^{4}$ \quad Weizhong Liu $^{1}$  \quad
	Chen Qian $^{4}$ \quad Ping Luo$^{2}$\\
	$^{1}$~Huazhong University of Science and Technology \quad
	$^{2}$~The University of Hong Kong \quad  \\
	$^{3}$~SenseTime Research \quad
	$^{4}$~SenseTime Research and Tetras.AI \\
	\tt\small jiahangwangchn@gmail.com \quad \{jinsheng, liuwentao, qianchen\}@sensetime.com \\
	\tt\small liuweizhong@mail.hust.edu.cn \quad pluo@cs.hku.hk 
}

\maketitle
\pagestyle{empty}
\thispagestyle{empty}
\begin{abstract}
Human pose estimation is a fundamental yet challenging task in computer vision, which aims at localizing human anatomical keypoints. However, unlike human vision that is robust to various data corruptions such as blur and pixelation, current pose estimators are easily confused by these corruptions. 
This work comprehensively studies and addresses this problem by building rigorous robust benchmarks, termed COCO-C, MPII-C, and OCHuman-C, to evaluate the weaknesses of current advanced pose estimators, and a new algorithm termed AdvMix is proposed to improve their robustness in different corruptions.
Our work has several unique benefits. (1) AdvMix is model-agnostic and capable in a wide-spectrum of pose estimation models. 
(2) AdvMix consists of adversarial augmentation and knowledge distillation. Adversarial augmentation contains two neural network modules that are trained jointly and competitively in an adversarial manner, where a generator network mixes different corrupted images to confuse a pose estimator, improving the robustness of the pose estimator by learning from harder samples. To compensate for the noise patterns by adversarial augmentation, knowledge distillation is applied to transfer clean pose structure knowledge to the target pose estimator. 
(3) Extensive experiments show that AdvMix significantly increases the robustness of pose estimations across a wide range of corruptions, while maintaining accuracy on clean data in various challenging benchmark datasets.

\def\thefootnote{$\dagger$}\footnotetext{The work was done during an internship at SenseTime Research.}\def\thefootnote{\arabic{footnote}}

\end{abstract}

\section{Introduction}

Human pose estimation (HPE) is a fundamental task for action recognition and video surveillance \cite{action_1, action_2, pose_surveilance}. Although convolutional neural networks (CNNs)  achieved great progress \cite{CPM, simple, hrnet, openpose, asso_embedding, higherhrnet} on challenging datasets \cite{coco, mpii, ochuman}, which only contain clean and high-resolution images, deploying models in the real world requires not only good performance on clean data, but also robustness to commonly occurring image corruptions. For example, while tracking and estimating the keypoints of a moving person in outdoor environments, current pose estimators suffer severe performance drop due to the noise or blur caused by weather conditions or camera systems. Therefore, analyzing and enhancing the robustness of pose estimators are important and are the purposes of this work.

\begin{figure}[t]
\centering
\includegraphics[width=0.45\textwidth]{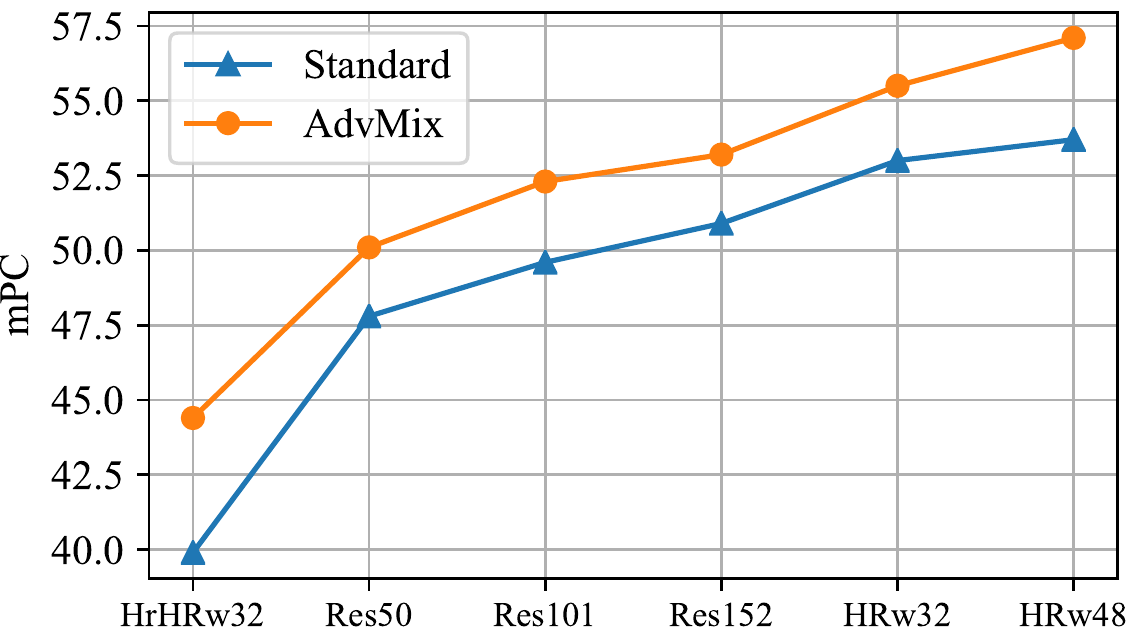}
\caption{\small \textbf{Improvements} of model robustness (mPC) when AdvMix is applied to the state-of-the-art methods.}
\label{fig:vis}
\end{figure}

Unlike previous studies on common robustness for classification, detection and segmentation \cite{imagenet_benchmark, det_benchmark, seg_benchmark}, human pose estimation uses a blend of classification and regression methods to model the structures of the human body, making it a challenging and collaborative field that is worthy of special investigations. The key challenges of robust human pose estimation are three folds. First, the lack of a benchmark for evaluating the robustness of state-of-the-art human pose estimation methods makes it difficult to construct rigorous comparisons between different models, not to mention to improve model robustness. Second, accuracy of clean data and corrupted data are trade-offs. Improving the robustness of the model while maintaining its performance on clean data is a non-trivial problem. Third, based on the proposed benchmark, we have examined the effectiveness of some data augmentation methods. However, we find that simply applying them sequentially does not achieve desirable performance. How can we effectively combine existing data augmentation techniques to improve the generalization of human pose estimators towards unforeseen corruptions? 

Inspired by~\cite{imagenet_benchmark}, we establish the robust pose benchmarks, consisting of three challenging datasets including COCO-C, MPII-C, and OCHuman-C. The benchmark datasets are constructed based on a full spectrum of \textbf{unforeseen} corruption types that are not encountered in model training (\ie CNNs are trained on \textit{clean} images, while evaluated on corrupted images). 
Extensive evaluations on these benchmarks show the weakness of both existing top-down and bottom-up pose estimators. (1) The state-of-the-art pose estimators suffer severe performance drop on corrupted images. (2) Models are generally more robust to brightness and weather changes, while less robust to motion and zoom blur. (3) The model robustness would increase by increasing model capacity. 

%
%

Empirical evaluations on the proposed benchmarks help us screen a collection of useful data augmentation techniques to improve model robustness under severe corruptions. In order to make full use of these techniques and achieve optimal performance on \textbf{unforeseen} noisy data, we propose an augmentation generator, which \textit{learns} to automatically combine augmented images. Specifically, we jointly train two neural networks in an adversarial manner, \ie an augmentation generator and a human pose estimator. The generator produces weights to mix up randomly augmented images, while the pose estimator attempts to learn robust visual representation from harder training samples.

It is worth noting that
%
the compositions produced by the augmentation generator may drift far from original images and such induced noise patterns may be harmful to performance on clean data. To reduce this negative impact, we propose to use a pre-trained teacher pose estimator to transfer structure knowledge learned from entire \textbf{clean} training data towards the target human pose estimator. Different from previous knowledge distillation methods that use a stronger network as the teacher model, our teacher pose estimator shares the same architecture as the target pose estimator. 
Extensive evaluations demonstrate that AdvMix significantly improves model robustness on diverse image corruptions while maintaining performance on clean data. The augmentation generator and the teacher pose estimator are only used for training and will be discarded at the inference stage, and thus introducing no computational overhead at inference time. Meanwhile, as shown in Fig.~\ref{fig:vis}, our method is model-agnostic and is proved to be effective for various state-of-the-art pose estimation models.

Our main \textbf{contributions} can be summarized as follows. 
\begin{itemize}
\item We propose three robust benchmarks COCO-C, MPII-C, and OCHuman-C, and demonstrate that both top-down and bottom-up pose estimators suffer severe performance drop on corrupted images, drawing the community's attention to this problem. 

\item With extensive experiments, we have many interesting conclusions that would help improve the accuracy and robustness of future works.

\item We propose a novel adversarial data augmentation method together with knowledge distillation, termed AdvMix, which is model-agnostic and easy-to-implement. It significantly improves the robustness of pose estimation models while maintaining or slightly improving the performance on the clean data, without extra inference computational overhead.

\end{itemize}

\section{Related Work}

\begin{figure*}[th]
\centering
\includegraphics[width=1.0\textwidth]{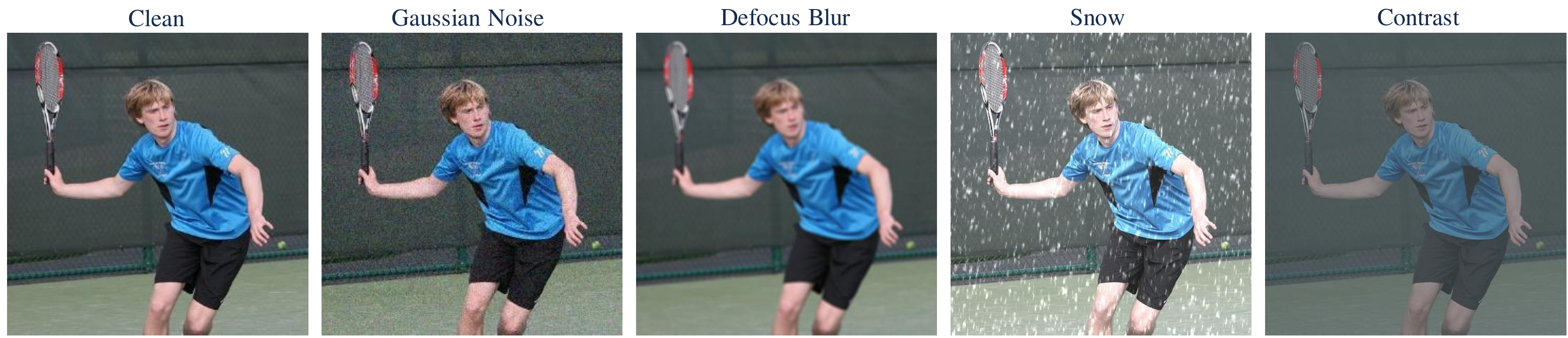}
\caption{\textbf{Visualization of examples in our benchmark datasets.} Our proposed benchmarks contain 15 different types of corruptions with different severity levels for a single clean image. The image corruption types are grouped into four main categories: noise, blur, weather, and digital. We sample one corruption type from each category in this figure.}
\label{corruption_images}
\end{figure*}

\subsection{Human Pose Estimation}
Human pose estimation (HPE) can be generally categorized into top-down and bottom-up methods. Top-down methods~\cite{CPM,stackhourglass,rmpe,hrnet,jin2020whole,liu2018cascaded, duan2019trb} divide the task into two stages: person detection and keypoint detection. SBL \cite{simple} presents a simple yet strong baseline network with several deconvolutional layers. HRNet \cite{hrnet} maintains high resolution representation combined with multi-scale feature fusions, and achieves the state-of-the-art performance on clean COCO dataset~\cite{coco}. 
Bottom-up methods~\cite{openpose,asso_embedding,pifpaf,higherhrnet,jin2020differentiable,jin2019multi} first detect all the keypoints and then group them into person instances. PifPaf \cite{pifpaf} utilized a Part Intensity Field (PIF) to localize body parts and a Part Association Field (PAF)~\cite{openpose} to associate body parts to form full human poses. HigherHRNet \cite{higherhrnet} learns scale-aware representations using high-resolution feature pyramids, and groups keypoints with associative embeddings~\cite{asso_embedding}. In this paper, we establish the benchmark and extensively evaluate the robustness of these state-of-the-art top-down and bottom-up methods.

\subsection{Corruption Robustness}
Recent studies have explored the corruption robustness of image classification \cite{dodge2016understanding,imagenet_benchmark}, object detection \cite{det_benchmark}, and segmentation \cite{seg_benchmark}. In comparison, the task of HPE is more comprehensive, which requires a blend of classification and regression approaches to model human body structure. Data denoising~\cite{buades2005review} \eg sparse and redundant representations~\cite{elad2006image}, non-local algorithm~\cite{buades2005non}, and denoising auto-encoder~\cite{xie2012image} are effective in removing noise. However, such methods are noise-specific, thus are not applicable to improving robustness towards unforeseen noises. To improve general robustness, recent works have explored several useful techniques such as pre-training~\cite{pretrain}, stability training~\cite{stability_training}, stylized image~\cite{imagenet_texture}, NoisyStudent~\cite{noisystudent}, and histogram equalization~\cite{imagenet_benchmark}.



\subsection{Data Augmentation}
Data augmentation has been widely utilized as an effective
method to improve model generalization. However, improving the general model robustness to unseen image corruptions is difficult. According to \cite{vasiljevic2016examining,generalization}, augmenting with one specific type of noise enhances the performance on the target noise, but it does not generalize to other unseen distortions while degrading the performance on clean data. Information dropping methods~\cite{random_erase,hide_and_seek,cutout,gridmask} and multi-image mixing methods~\cite{mixup,cutmix,augmix} have gained decent improvements on clean data for image classification. Learned augmentation methods~\cite{adv_autoaug, autoaugment} have also been proposed to improve performance. For pose estimation, adversarial data augmentation methods~\cite{advpose_1, advpose_2} are leveraged to optimize augmentation hyper-parameters, \eg the rotation angle. However, they only focus on searching for common augmentation hyper-parameters and simply combine different augmentations sequentially to improve performance on clean data. Instead, we adversarially learn attention weights for mixing randomly augmented images to enhance model robustness. AugMix \cite{augmix} is proposed to mix augmented images using beta or dirichlet coefficients to boost the robustness on image classification. Instead of simply using a fixed augmentation sampler(\eg dirichlet distribution) to generate mixing weights, our AdvMix uses adversarial training to learn to generate appropriate mixing attention weights for each training sample. 

\section{Methods}
\subsection{Robust Pose Benchmark}

\begin{figure*}[t]
\centering
\includegraphics[width=1.0\textwidth]{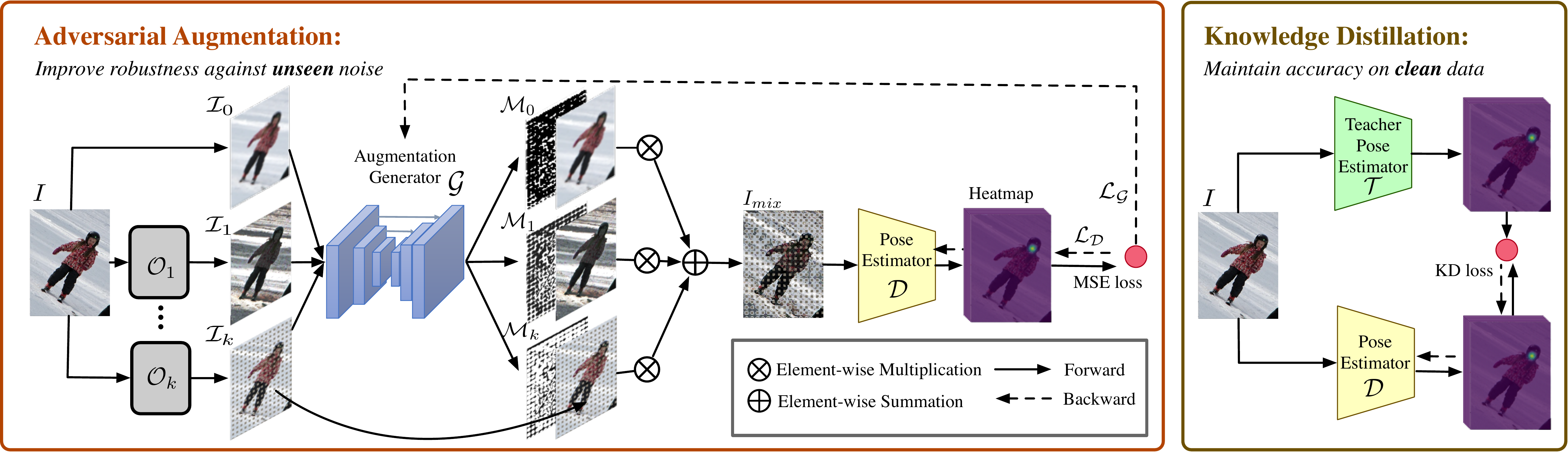}
\caption{\textbf{Overview of AdvMix.} Our framework consists of two modules: the adversarial augmentation module and the knowledge distillation module. The adversarial augmentation module contains an augmentation generator and a pose estimator. For example, given two (\ie $K=2$) differently augmented images, the augmentation generator estimates the corresponding attention maps $\mathcal{M}$, and mixes them up to the final image $I_{mix}$, while the pose estimator generates keypoint heatmaps. They are trained in an adversarial manner following Eq.(\ref{optimizaton_process}). The robustness of the pose estimator can be significantly improved if the generator cannot confuse it. To reduce the effect of induced noise patterns, we use a teacher pose estimator for transferring pose structure knowledge in adversarial training. The teacher pose estimator shares the same architecture as the target pose estimator and is pre-trained on the entire clean data.} 
\label{fig:framework}
\end{figure*}

\subsubsection{Benchmark Datasets}
The robust pose estimation benchmark is composed of three benchmark datasets: COCO-C, MPII-C, OCHuman-C, which are constructed by applying 15 different types of image corruptions~\cite{imagenet_benchmark} with 5 severity levels to the official \textit{validation} set of COCO~\cite{coco}, MPII~\cite{mpii} and OCHuman~\cite{ochuman}. Therefore, the total number of each benchmarking dataset is $15 \times 5$ times that of the corresponding validation dataset. The types of image corruption are sorted into four main categories (noise, blur, weather, and digital), which are diverse and enormous enough to cover real-world corruptions. 

\textbf{COCO-C Dataset.} COCO-C dataset is constructed from COCO~\cite{coco} val2017 set. The COCO dataset contains over 200,000 images and 250,000 person instances, where there are 5000 images for the val2017 set.

\textbf{MPII-C Dataset.} MPII-C dataset is constructed from MPII~\cite{mpii} test set. The MPII dataset consists of images taken from a wide range of real-world activities with full-body pose annotations. There are around 25K images with 40K subjects, where there are 12K subjects for testing and the remaining subjects for the training set.

\textbf{OCHuman-C Dataset.} OCHuman-C is constructed from OCHuman~\cite{ochuman}. The OCHuman dataset focuses on heavily occluded human with comprehensive annotations including bounding-box, humans pose and instance mask, which contains 13360 elaborately annotated human instances within 5081 images.

\subsubsection{Evaluation Metrics}
For COCO~\cite{coco} and OCHuman~\cite{ochuman}, the standard average precision (AP) is used to evaluate the models. In this paper, we use $\text{AP}^*$ to denote the performance on \textbf{clean} data. To evaluate model robustness, we follow~\cite{det_benchmark} to use mean performance under corruption ($\text{mPC}$):

\begin{equation}
    \text{mPC} = \frac{1}{N_c} \sum\limits_{c=1}^{N_c} \frac{1}{N_s} \sum\limits_{s=1}^{N_s} \text{AP}_{c,s}.
    \label{mPC}
\end{equation}
Here, $\text{AP}_{c,s}$ is the average performance measured on corruption type $c$ under severity level $s$. $N_c=15$ and $N_s=5$ are the numbers of corruption types and severity levels, respectively. To assess the robustness of models, we define the relative performance under corruption ($\text{rPC}$) as follows:

\begin{equation}
    \text{rPC} = \frac{\text{mPC}}{\text{AP}^*}.
    \label{rAP}
\end{equation}

For MPII~\cite{mpii} dataset, the official evaluation metric is PCKh. Similarly, we use $\text{PCKh}^*$ to denote the performance on clean data. Similar to mPC and rPC in Eq.(\ref{mPC},\ref{rAP}), mean performance under corruption ($\text{mPC}$) and relative mean performance under corruption ($\text{rPC}$) are introduced as follows:

\begin{equation}
    \text{mPC} = \frac{1}{N_c} \sum\limits_{c=1}^{N_c} \frac{1}{N_s} \sum\limits_{s=1}^{N_s} \text{PCKh}_{c,s}
    \label{mPC_mpii}
\end{equation}

\begin{equation}
    \text{rPC} = \frac{\text{mPC}}{\text{PCKh}^*}.
    \label{rPCKh}
\end{equation}

\subsection{Adversarial Augmentation Mix (AdvMix)}




\subsubsection{Augmentation Generator}

As shown in Figure \ref{fig:framework}, given an image $I$, we first randomly generate $K$ differently augmented images $\mathcal O_k(I)$ using parallel augmentation strategies $\mathcal O_k$. We use $K=2$ by default in our implementation. Together with the original image $I$, we get a set of $K + 1$ proposal images. The augmentation generator $\mathcal{G}(\cdot, \theta)$ is applied to output the normalized attention maps $\mathcal{M}^{(K + 1) \times H \times W}$, where $H$ and $W$ are the image resolution. The attention maps $\mathcal{M}$ are used as the weights to mix up the proposal images, following Eq.(\ref{eq:mixed_image}), where $\odot$ is the Hadamard product.

\begin{equation}
    I_{mix} =\mathcal{M}_0 \odot {I} + \sum_{k=1}^{K}{\mathcal{M}_k \odot \mathcal O_k(I)}.
    \label{eq:mixed_image}
\end{equation}

Based on the benchmark (Table.\ref{tab:benchmark_all}), we examine and find a collection of useful techniques to improve the model robustness for pose estimation. Grid-Mask \cite{gridmask} applies information removal to achieve state-of-the-art results in various computer vision tasks. AutoAugment~\cite{autoaugment} automatically searches for a mixture of augmentation policies. It improves both the clean performance and the robustness for image classification~\cite{fourier}. To avoid over-fitting to the test set of held-out corruptions, we manually exclude some operations, such as contrast, color, brightness, and sharpness sub-policies (as they appear in the benchmark). Given an image $I$, we choose Grid-Mask and AutoAugment to generate randomly augmented images $\mathcal O$, and then mix these images through mixing weight from augmentation generator. We adopt the U-Net~\cite{Unet} architecture to build the augmentation generator $\mathcal{G}(\cdot, \theta)$, which generates the attention maps for mixing up randomly augmented images. Please refer to the supplementary for more implementation details.



\subsubsection{Heatmap Regression}

The human body pose is encoded with 2D Gaussian confidence heatmaps, where each channel corresponds to one body keypoint. The objective of human pose estimation network $\mathcal{D}(\cdot, \phi)$ is to minimize the MSE loss between the predicted heatmaps and the ground truth heatmaps $\mathcal{H}_{gt}$:

\begin{equation}
    \mathcal{L}_\mathcal{D^*} = \left\| \mathcal{D}(I_{mix}, \phi) - \mathcal{H}_{gt}\right\|_2^2.
    \label{eq:mse_loss}
\end{equation}

To enhance the supervision to the target pose estimator $\mathcal{D}$, we introduce a teacher pose estimator $\mathcal{T}$ for knowledge distillation and providing softened heatmap labels. Note that $\mathcal{T}$ and $\mathcal{D}$ share the same architecture and $\mathcal{T}$ is pre-trained on the clean training data. The parameters of the $\mathcal{T}$ will be fixed while training. We select MSE loss for knowledge distillation loss as:

\begin{equation}
    \mathcal{L}_{\mathcal{D}_{kd}}= \left\| \mathcal{D}(I, \phi) - \mathcal{T}(I)\right\|_2^2.
    \label{eq:kd_loss}
\end{equation}

We formulate the overall loss function of pose estimator $\mathcal{D}$ while training as:

\begin{equation}
    \mathcal{L}_\mathcal{D} = (1 - \alpha) \mathcal{L}_\mathcal{D^*} + \alpha \mathcal{L}_{\mathcal{D}_{kd}}.
    \label{eq:overall_D_loss}
\end{equation}

where $\alpha$ is the loss weight balancing between MSE loss and knowledge distillation loss. We observe that our model is not sensitive to $\alpha$. Thus we choose $\alpha=0.1$ as the default setting in our experiments.

\subsubsection{Adversarial Training}

The augmentation generator $\mathcal{G}(\cdot, \theta)$ and human pose estimator $\mathcal{D}(\cdot, \phi)$ are trained in an adversarial manner. $\mathcal{G}(\cdot, \theta)$ tries to find the most confusing way to mix up randomly augmented images, while $\mathcal{D}(\cdot, \phi)$ learns more robust features from harder training samples. The optimization objective of augmentation generator is defined as:

\begin{equation}
    \mathcal{L}_{\mathcal{G}} = -\mathcal{L}_{\mathcal{D^*}}.
    \label{eq:G_loss}
\end{equation}

Overall, the whole learning process can be defined as two-player zero-sum game with value function $\mathcal{V}(\mathcal{D}, \mathcal{G})$. 
%

\begin{equation}
    \mathcal{V}(\mathcal{D}, \mathcal{G}) = \mathop{\min}_{\phi} \mathop{\max}_{\theta}\mathop{\mathbb{E}}_{\mathcal{I}\sim\Omega} \mathcal{L}(\mathcal{D}(\mathcal{G}(\mathcal{\mathcal{O}(\mathcal{I}), \theta)}, \phi),\mathcal{H}_{gt}).
    \label{optimizaton_process}
\end{equation}

\begin{table*}[tb]
    \centering
    \caption{\small Pose robustness benchmark for both top-down and bottom-up models on COCO-C, MPII-C and OChuman-C. AP$^*$ and PCKh$^*$ represent the performance on clean data. mPC represents mean performance under all corruptions, while rPC measures the relative performance. The remaining columns are the breakdown APs for different corruptions. We see that performances of existing advanced pose estimators significantly drop when corruptions are presented.}
    \scalebox{0.59}{
    \begin{tabular}{c|c|c|c|a|a|ccc|cccc|cccc|cccc}
\hline
\multicolumn{21}{c}{} \\[-5pt]
\multicolumn{21}{c}{\large Results of top-down methods on COCO-C with the same detection bounding boxes as \cite{hrnet}} \\[5pt]
\hline
Method & Backbone & Input size & AP$^*$ & mPC & rPC & Gauss& Shot & Impulse & Defoucs & Glass & Motion & Zoom & Snow & Frost & Fog & Bright & Contrast & Elastic & Pixel & JPEG \\
\hline
\multirow{6}*{SBL~\cite{simple}}
&\multirow{3}*{ResNet-50}
& $128 \times 96$ &
59.0&40.7&69.0&37.8&39.8&36.9&39.5&38.1&35.3&13.3&39.1&43.0&48.1&54.5&38.5&49.2&50.9&47.1\\
&& $256 \times 192$ & 70.4&47.8&67.9&45.8&48.1&45.6&43.4&42.1&38.8&16.3&49.1&52.5&58.9&65.5&47.2&56.7&55.2&51.7\\
&& $384 \times 288$ &72.2&47.7&66.1&45.2&47.8&45.9&42.9&42.1&38.1&16.3&49.9&53.2&60.0&66.8&48.2&57.3&53.2&49.2\\
&\multirow{3}*{ResNet-101}
& $128 \times 96$&
61.1&42.6&69.7&40.4&42.6&39.6&40.7&39.9&36.8&14.5&41.3&45.2&49.6&56.8&38.6&51.1&53.2&49.1\\
&& $256 \times 192$ & 71.4&49.6&69.5&47.8&50.1&47.2&45.1&43.8&40.3&17.6&50.9&54.9&60.9&67.0&49.7&58.1&57.0&53.8 \\
&& $384 \times 288$&
73.6&50.4&68.4&49.2&51.7&49.1&44.8&43.9&40.0&17.7&52.5&56.4&62.7&69.0&51.2&59.0&55.5&52.7\\
\hline
\multirow{6}*{HRNet~\cite{hrnet}}
&\multirow{3}*{HRNet-W32}
& $128 \times 96$&
66.9&47.2&70.6&42.7&45.7&43.0&44.0&43.1&40.9&15.9&47.1&51.4&57.4&63.0&47.5&55.7&57.2&53.4 \\
&& $256 \times 192$ &74.4&53.0&71.3&51.3&54.2&52.6&46.9&46.3&43.5&19.2&55.9&59.1&65.2&70.3&54.1&60.5&59.4&56.9  \\
&& $384 \times 288$ & 75.7&53.7&70.9&51.9&54.7&53.7&47.8&47.1&43.8&19.8&57.9&60.3&66.5&71.6&55.4&61.1&58.1&55.8 \\
&\multirow{3}*{HRNet-W48}
& $128 \times 96$ &
68.6&49.3&71.8&45.9&48.8&46.1&45.3&44.4&42.4&16.5&49.6&54.1&59.8&65.0&49.4&57.3&59.2&55.2
\\
&& $256 \times 192$ &75.1&53.7&71.6&52.5&55.2&53.4&46.8&46.7&43.5&19.1&57.0&60.1&66.4&71.4&55.2&61.1&60.0&57.6\\
&& $384 \times 288$ & 76.3&54.2&71.1&52.8&55.8&54.2&47.6&47.3&43.4&19.5&58.3&60.9&67.5&72.3&56.3&61.6&59.2&57.1 \\
\hline

\multicolumn{21}{c}{} \\[-5pt]
\multicolumn{21}{c}{\large Results of top-down methods on MPII-C} \\[5pt]
\hline
Method& Backbone & Input size & PCKh$^*$ & mPC & rPC & Gauss& Shot & Impulse & Defoucs & Glass & Motion & Zoom & Snow & Frost & Fog & Bright & Contrast & Elastic & Pixel & JPEG \\
\hline
\multirow{3}*{SBL~\cite{simple}}
&Resnet-50 & $256 \times 256$ & 88.5&77.5&87.6&68.9&71.6&68.9&84.2&85.3&83.6&53.7&70.0&75.6&83.1&86.6&69.1&87.8&87.8&86.8 \\
&Resnet-101 & $256 \times 256$ & 
89.1&78.6&88.3&70.4&73.1&70.6&84.8&86.0&84.1&53.9&72.2&76.9&84.1&87.0&71.9&88.3&88.4&87.3\\
&Resnet-152&$256 \times 256$ &
89.6&79.6&88.8&74.1&76.3&74.1&85.3&86.5&84.8&54.5&72.1&77.6&84.4&87.7&71.4&88.8&88.8&87.9\\
\hline
HRNet~\cite{hrnet}&HRNet-W32& $256 \times 256$ & 90.3&80.1&88.7&71.5&74.1&72.4&86.0&87.0&85.3&56.0&73.8&78.6&86.1&88.5&74.5&89.5&89.5&88.7 \\
\hline
\multicolumn{21}{c}{} \\[-5pt]
\multicolumn{21}{c}{\large Results of bottom-up methods on COCO-C} \\[5pt]
\hline
Method & Backbone & Input size & AP$^*$ & mPC & rPC & Gauss& Shot & Impulse & Defoucs & Glass & Motion & Zoom & Snow & Frost & Fog & Bright & Contrast & Elastic & Pixel & JPEG \\
\hline
\multirow{3}*{PifPaf~\cite{pifpaf}}
& ShuffleNet V2 & $641 \times 641$ &60.7&32.9&54.2&25.7&27.3&24.6&29.1&28.5&26.4&8.8&33.5&38.5&47.6&54.3&34.1&45.7&33.4&35.7\\
& Resnet-50 & $641 \times 641$ &64.8&34.4&53.1&30.4&32.2&29.5&30.9&27.9&26.5&8.8&36.4&40.2&49.9&57.3&34.9&47.8&29.9&34.1\\
& Resnet-101 & $641 \times 641$ & 68.3&40.6&59.5&38.0&39.4&36.6&36.9&34.6&31.6&11.2&42.6&46.6&56.0&62.0&42.4&52.6&37.2&41.3 \\
\hline
\multirow{4}*{HrHRNet~\cite{higherhrnet}}
& \multirow{2}*{HrHRNet-W32}
& $512 \times 512$ & 67.1&39.9&59.4&34.2&37.0&35.2&35.1&32.5&34.0&12.5&43.3&47.6&54.9&60.6&43.4&50.3&42.2&35.0 \\
& & $640 \times 640$ &68.5&39.6&57.8&31.5&38.9&37.6&34.8&32.6&33.8&12.1&44.0&47.9&55.1&61.4&42.9&50.6&37.8&33.2\\
& \multirow{2}*{HrHRNet-W48}
&$512 \times 512$ & 68.5&41.9&61.2&39.3&42.2&40.1&36.1&33.5&35.1&13.1&45.0&49.1&56.7&62.1&45.5&51.4&42.2&36.9 \\
& &$640 \times 640$ &69.8&40.8&58.4&36.7&39.8&37.9&35.6&33.1&34.2&12.5&44.7&49.0&56.3&62.8&43.7&51.3&39.9&34.3 \\
\hline
\multicolumn{21}{c}{} \\[-5pt]
\multicolumn{21}{c}{\large Results of bottom-up methods on OCHuman-C} \\[5pt]
\hline
Method & Backbone & Input size & AP$^*$ & mPC & rPC & Gauss& Shot & Impulse & Defoucs & Glass & Motion & Zoom & Snow & Frost & Fog & Bright & Contrast & Elastic & Pixel & JPEG \\
\hline
\multirow{3}*{PifPaf~\cite{pifpaf}}
& ShuffleNet V2 & $641 \times 641$ & 37.8&30.8&81.5&28.7&28.8&28.1&32.7&32.2&32.9&17.2&26.3&30.5&36.4&36.1&30.5&36.2&32.8&32.6\\
& Resnet-50 & $641 \times 641$ & 37.6&30.8&82.0&30.1&30.0&29.4&33.3&31.6&32.4&16.5&27.0&30.5&36.5&36.5&29.9&36.1&31.8&30.7\\
& Resnet-101 & $641 \times 641$ & 39.6&34.9&88.1&35.6&35.6&35.3&37.2&35.3&36.3&20.2&31.4&34.4&39.2&38.8&34.8&38.9&35.5&34.6 \\
\hline
\multirow{4}*{HrHRNet~\cite{higherhrnet}}
& \multirow{2}*{HrHRNet-W32}
& $512 \times 512$ & 40.0&35.1&87.6&32.9&33.1&33.2&37.3&36.5&36.9&21.7&31.6&36.0&39.9&39.4&36.9&39.2&37.1&34.4 \\
& & $640 \times 640$ & 39.3&33.6&85.4&30.9&31.1&31.8&36.2&35.0&36.1&19.6&31.1&35.1&39.3&38.1&35.8&38.3&33.8&31.1\\
& \multirow{2}*{HrHRNet-W48}
& $512 \times 512$ & 
41.7&36.7&87.9&35.5&35.6&35.7&38.8&37.5&38.5&22.0&33.5&37.6&41.5&40.5&39.0&40.6&38.0&35.6 \\
& & $640 \times 640$ & 40.9&35.4&86.4&33.0&33.3&33.5&37.7&36.6&37.6&21.1&32.4&37.4&41.5&39.9&37.2&39.9&36.6&32.6\\
\hline
    \end{tabular}}
    \label{tab:benchmark_all}
\end{table*}




\section{Evaluation on Robust Pose Benchmark}
\subsection{Experimental Setup}
We extensively evaluate the performance of the state-of-the-art methods on the proposed Robust Human Pose Benchmark. To assess the robustness of human pose estimators, the models are only trained on clean data (\eg COCO) and then evaluated on corrupted data (\eg COCO-C). Note that OCHuman dataset is only designed for validation, we follow the common settings~\cite{ochuman} to train the models on COCO and evaluate on OCHuman-C. For top-down methods (\ie\textbf{SBL}~\cite{simple}, \textbf{HRNet}~\cite{hrnet}) with input size $256 \times 192$ and $384 \times 288$, we use the officially trained checkpoints~\footnote{https://github.com/leoxiaobin/deep-high-resolution-net.pytorch}, and then directly evaluate their robustness performance on on COCO-C and MPII-C datasets. Since the pre-trained models with image size $128 \times 96$ are not publicly available, we retrain these models following the official training settings. For bottom-up methods (\ie\textbf{PifPaf}~\cite{pifpaf}, \textbf{HigherHRNet}~\cite{higherhrnet}), we also follow the official codes and settings. The pre-trained models of PifPaf~\footnote{https://github.com/vita-epfl/openpifpaf/tree/v0.10.0} and HigherHRNet~\footnote{https://github.com/HRNet/HigherHRNet-Human-Pose-Estimation} are directly used for evaluating the robustness using on COCO-C and OCHuman-C. The results are reported in Table \ref{tab:benchmark_all}.

\subsection{Benchmarking Conclusions}

\subsubsection{Pose estimation methods performance}
\begin{figure*}[t]
\centering
\includegraphics[width=0.95\textwidth]{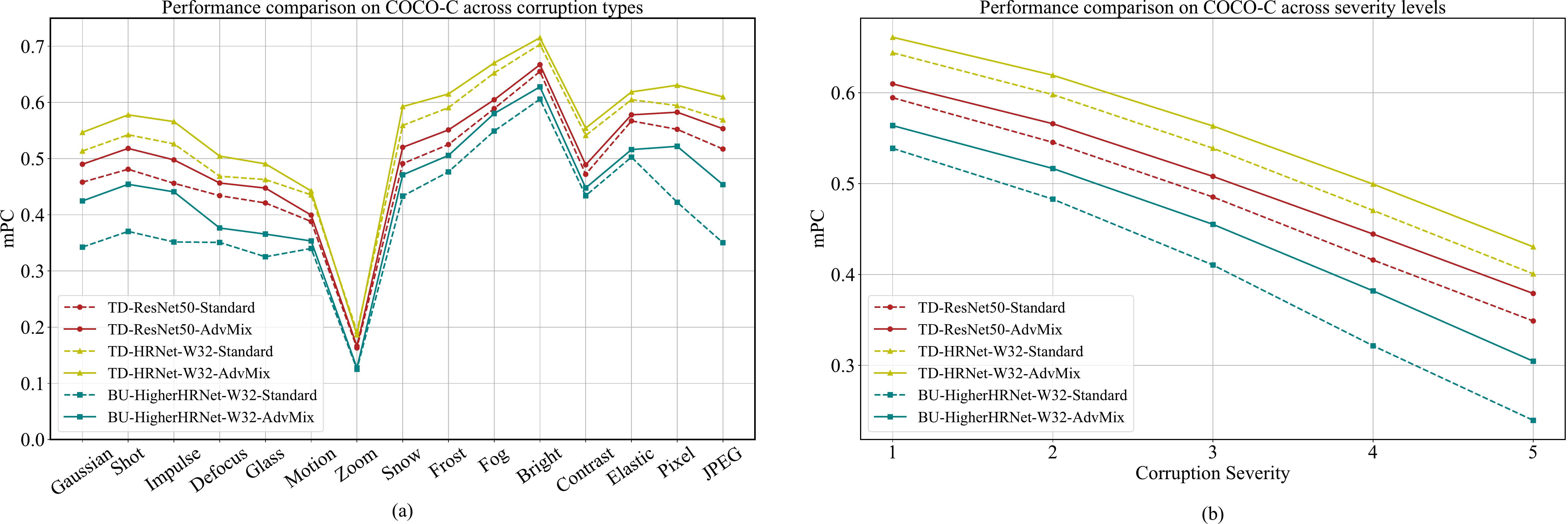}
\caption{(a) Performance improvements of AdvMix on COCO-C across different corruption types. (b) Performance improvements of AdvMix on COCO-C across different corruption severities. The results are obtained with input sizes as $256 \times 192$. Input sizes for top-down (TD) methods are $256 \times 192$, while for bottom-up (BU) methods (\eg HigherHRNet) is $512 \times 512$.}
\label{fig:method_robustness}
\end{figure*}

\textbf{Top-down and bottom-up models show similar performance degradation tendency on different corruptions.} As shown in Figure \ref{fig:method_robustness}, the model robustness to different types of corruption varies a lot. However, performance degradation across different models are similar. For example, all models are more robust to weather or brightness changes, while less robust to motion or zoom blur. 

\begin{figure*}[t]
\centering
\includegraphics[width=0.90\textwidth]{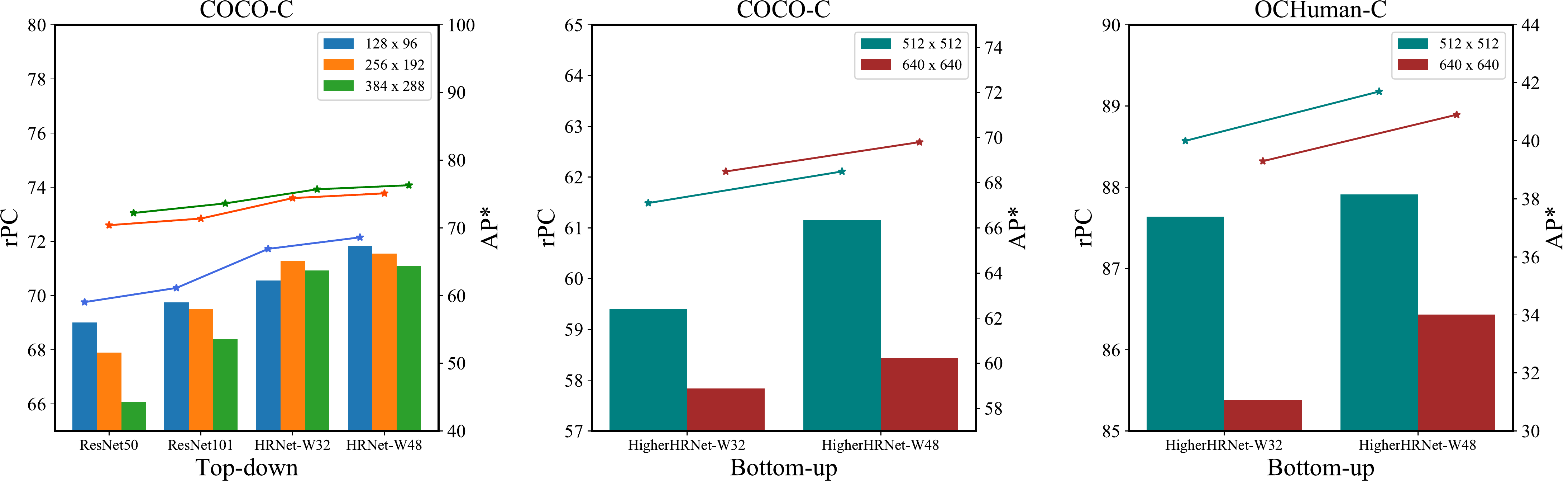}
\caption{Performance of top-down and bottom-up models with different backbones and input sizes on COCO-C and OCHuman-C. The line graph shows performance on clean data (AP$^*$), and the bar graph denotes robustness score (rPC). }
\label{fig:arch_robustness}
\end{figure*}

\subsubsection{Backbone network performance}
\textbf{Robustness increases with model capacity.} With the same pose estimation methods, the robustness of different backbone networks varies a lot and increases with model capacity. 
In Figure~\ref{fig:arch_robustness}, the line graph shows the performance on clean data (AP$^*$), while the bar graph shows the robustness score (rPC). The color denotes different input image resolutions. We observe that with the same input resolution, a model with higher capacity (HRNet) is generally more accurate and more robust than that with lower capacity (ResNet). This indicates that robustness can be improved by using a stronger backbone network.
%


\section{Robustness Enhancement with AdvMix}

\begin{table}[tbh]
    \centering
    \caption{\small \textbf{Comparisons} between standard training and AdvMix on COCO-C. For top-down approaches, results are obtained with detected bounding boxes of \cite{hrnet}. We see that mPC and rPC are greatly improved, whilst clean performance AP$^*$ can be preserved.}
    \scalebox{0.8}{
    \begin{tabular}{c|c|c|c|c|c}
\hline
Method & Backbone &  Input size &AP$^*$ & mPC & rPC\\
\hline
Standard &ResNet-50 & $256 \times 192$
&\textbf{70.4} & 47.8 &67.9 \\
\textbf{AdvMix}&ResNet-50 & $256 \times 192$
&70.1 & \textbf{50.1} & \textbf{71.5}\\  \hline
Standard &ResNet-101 & $256 \times 192$
&\textbf{71.4} & 49.6 & 69.5 \\
\textbf{AdvMix}&ResNet-101 & $256 \times 192$
&71.3 & \textbf{52.3} & \textbf{73.3}\\ \hline
Standard &ResNet-152 & $256 \times 192$
& 72.0 & 50.9 &  70.7 \\
\textbf{AdvMix}&ResNet-152 & $256 \times 192$
& \textbf{72.3} & \textbf{53.2} & \textbf{73.6}\\
\hline
Standard&HRNet-W32 & $256 \times 192$
&74.4 & 53.0 & 71.3\\
\textbf{AdvMix}&HRNet-W32& $256 \times 192$
&\textbf{74.7}&\textbf{55.5}&\textbf{74.3} \\ \hline
Standard&HRNet-W48 & $256 \times 192$
& 75.1 & 53.7 & 71.6 \\
\textbf{AdvMix}&HRNet-W48& $256 \times 192$
&\textbf{75.4}&\textbf{57.1}&\textbf{75.7} \\ \hline \hline
Standard&HrHRNet-W32 & $512 \times 512$
&67.1 &39.9	&59.4\\
\textbf{AdvMix}&HrHRNet-W32& $512 \times 512$ 
&\textbf{68.3} &\textbf{45.4}	&\textbf{66.5} \\
\hline
    \end{tabular}}
    \label{tab:advmix_coco}
\end{table}

\begin{table}[tbh]
    \centering
    \caption{\small \textbf{Comparisons} between standard training and AdvMix of representative top-down methods on MPII-C. We draw similar conclusion that AdvMix is effective to increase mPC and rPC, whilst maintaining clean performance PCKh$^*$.}
    \scalebox{0.8}{
    \begin{tabular}{c|c|c|c|c|c}
\hline
Method & Backbone & Input size & PCKh$^*$ & mPC & rPC\\
\hline
Standard&ResNet-50 &$256 \times 256$
&88.5 &77.5	&87.6 \\
\textbf{AdvMix}&ResNet-50&$256 \times 256$
& \textbf{88.9} & \textbf{82.0} & \textbf{92.3}\\\hline
Standard&ResNet-101 &$256 \times 256$
&89.1 &78.6	& 88.3 \\
\textbf{AdvMix}&ResNet-101&$256 \times 256$
& \textbf{89.4} & \textbf{82.8} & \textbf{92.5}\\
\hline
Standard&HRNet-W32& $256 \times 256$
& 90.3 &80.1&88.7\\
\textbf{AdvMix}&HRNet-W32 &$256 \times 256$
&\textbf{90.5} &\textbf{83.9} & \textbf{92.7} \\ \hline
    \end{tabular}}
    \label{tab:advmix_mpii}
\end{table}

\subsection{Implementation Details}
For adversarial training, we simply reuse the existing publicly available well-trained checkpoints and retrain the models with AdvMix on clean data. We simply follow the same training settings (\ie the same hyper-parameters, learning rate, and total training epochs) as the official codes. By default, the initial learning rates for the augmentation generator and human pose estimator is 0.001. We decay the learning rate by the factor of 10 at the 170-th epoch, and 200-th epoch. The adversarial training ends at 210-th epoch. We adopt ADAM optimizer to train both the augmentation generator and the human pose estimator. 
All experiments are conducted using PyTorch~\cite{paszke2017automatic} on NVIDIA TITAN X Pascal GPUs.

\begin{table*}[h]
    \centering
    \caption{\small Results of directly augmenting with the test-time corruption types (Transfer). We use $^\dagger$ to denote the selected corruption types (\eg Gaussian noise, Defocus blur, Snow, and Contrast) are used as data augmentation during training. }
    \scalebox{0.69}{
    \begin{tabular}{c|c|c|ccc|cccc|cccc|cccc}
\hline
Method & AP$^*$ & mPC & Gauss& Shot & Impulse & Defoucs & Glass & Motion & Zoom & Snow & Frost & Fog & Bright & Contrast & Elastic & Pixel & JPEG \\
\hline
Standard&76.6&53.3&51.1&54.1&52.5&46.7&46.2&43.4&18.8&56.2&59.7&66.2&71.8&54.5&61.1&59.8&57.1 \\
Transfer&\textcolor{gray}{75.4}&56.2&58.7$^\dagger$&60.3&59.7&55.0$^\dagger$&48.3&44.3&19.0&65.6$^\dagger$&61.7&\textcolor{gray}{65.8}&\textcolor{gray}{70.9}&58.9$^\dagger$&\textcolor{gray}{60.3}&61.8&\textcolor{gray}{52.2} \\
\hline
AdvMix&\textbf{77.1}&55.9&54.7&57.9&56.7&50.4&49.1&44.2&\textcolor{gray}{18.2}&59.8&62.4&68.2&73.2&56.0&62.6&63.4&61.5 \\
AdvMix+Stylized&76.5&\textbf{56.5}&54.5&57.5&56.4&50.4&49.3&46.1&20.0&62.3&63.0&67.6&72.9&59.0&62.5&63.9&61.8 \\
\hline
    \end{tabular}}
    \label{tab:type}
\end{table*}

\begin{table}[h]
    \centering
    \caption{\small \textbf{Comparisons} with other techniques on COCO-C with HRNet-W32 backbone. The results are obtained using ground truth bounding boxes. We can observe that AdvMix significantly outperforms the state-of-the-art methods on robustness, while preserving the performance of clean AP.}
    \scalebox{0.8}{
    \begin{tabular}{c|c|c|c}
\hline
Method & AP$^*$ & mPC & rPC\\
\hline
Standard&
76.6&53.3&69.6\\
FPD~\cite{zhang2019fast}&
77.3&54.0&69.9\\
FPD$^\dagger$~\cite{zhang2019fast}&
76.6&53.4&69.7\\
Grid-Mask~\cite{gridmask}&
76.4&54.8&71.7\\
AutoAugment~\cite{autoaugment}&
76.2&54.2&71.1\\
Stylized Only~\cite{imagenet_benchmark,det_benchmark}&
67.5& 46.7 & 69.1\\
Stylized~\cite{imagenet_benchmark,det_benchmark}&
76.1&54.8&72.0\\ \hline
\textbf{AdvMix} &
\textbf{77.1}& 55.9& 72.5\\
Stylized + \textbf{AdvMix}&
76.5& \textbf{56.5}&\textbf{73.8}\\
\hline
    \end{tabular}}
    \label{tab:method_comparison}
\end{table}

\begin{table}[h]
    \centering
    \caption{\small \textbf{Ablation} study on COCO-C with HRNet-W32 backbone. The results are obtained using ground truth bounding boxes. We see that AdvMix performs better than other augmentation composing approaches on robustness by large margin.}
    \scalebox{0.8}{
    \begin{tabular}{c|c|c|c}
\hline
Method & AP$^*$ & mPC & rPC\\
\hline
SequentialMix &
76.2&54.6&71.6\\
EqualMix& 76.6 & 54.2 & 70.7 \\
DirichletMix~\cite{augmix}& 76.5 & 54.4 & 71.1 \\ \hline
\textbf{AdvMix}-Image & \textbf{77.5}& 55.1& 71.2\\
\textbf{AdvMix} &
77.1& \textbf{55.9}& \textbf{72.5} \\
\hline
    \end{tabular}}
    \label{tab:ablation}
\end{table}

\begin{table}[h]
    \centering
    \caption{Effect of knowledge distillation.}
    \scalebox{0.8}{
    \begin{tabular}{c|c|c|c|c}
\hline
Method & dataset & AP$^*$/ PCKh$^*$ & mPC & rPC\\
\hline
\textbf{AdvMix} w/o KD  & COCO-C & 76.8 & 55.9 & 72.7 \\
\textbf{AdvMix} & COCO-C & 
\textbf{77.1}& 55.9 & 72.5 \\
\hline
\textbf{AdvMix} w/o KD  & MPII-C & 89.9 & 81.7 & 90.9 \\
\textbf{AdvMix} & MPII-C &  
\textbf{90.5} &\textbf{83.9} & \textbf{92.7} \\
\hline
    \end{tabular}}
    \label{tab:ablation_kd}
\end{table}


\subsection{Quantitative Results}
As shown in Table~\ref{tab:advmix_coco} and Table~\ref{tab:advmix_mpii}, we find that the proposed method significantly improves the model robustness while the performance on clean data is maintained or slightly improved. We also observe that AdvMix significantly boosts the robustness performance for bottom-up methods ($39.9$ to $45.4$ mPC) and the robustness gain is larger than that of top-down methods shown in Table~\ref{tab:advmix_coco}. 
Figure~\ref{fig:method_robustness}(a) shows the performance degradation and the robustness improvement of different models across different corruptions. Performance degradation among different models shows a similar tendency. Models are generally more robust to brightness and weather changes, while less robust to motion and zoom blur. AdvMix performs better than baseline by different corruption types and the gain of noise and digital distortion are larger than other corruption types.
Comparing the performance across different corruption severity in Figure~\ref{fig:method_robustness}(b), we observe that AdvMix consistently improves over the baseline, and the gain gets larger for severer corruption.

We compare AdvMix with other state-of-the-art methods. FPD~\cite{zhang2019fast} proposes to improve the performance of lightweight models through knowledge distillation, and the teacher net is more sophisticated than the student net. FPD$^\dagger$ is our implementation with the same setting as FPD except that the teacher net uses the same architecture as the student net. Stylized dataset~\cite{style_transfer} is also used as a technique to improve robustness. As shown in Table~\ref{tab:method_comparison}, we can see that though FPD can boost the performance on clean data, the improvement of robustness is limited. 
We also find that only using knowledge distillation (teacher use the same architecture in FPD$^\dagger$) without AdvMix will not improve the performance on clean data and model robustness.
AdvMix can not only improve the mPC, but also slightly improve the performance on clean data, outperforming operating AutoAugment or Grid-Mask separately. 
Finally, we also prove that AdvMix can be combined with the data stylizing technique and further enhance the robustness on corrupted images while almost maintaining the performance of clean data.

\subsection{Ablation Studies}
\label{sec:exp_ablation}

We conduct ablation studies on COCO-C dataset using HRNet-W32 backbone with input size $256 \times 192$ to validate the effectiveness of the augmentation composition method in AdvMix. 
SequentialMix simply composes augmentation operations in a chain and applies them sequentially. EqualMix mixes the augmented images from various augmentation chains using equal weights. DirichletMix follows \cite{augmix} to sample the mix weights from Dirichlet distribution to mix the augmented images. In comparison, AdvMix uses the augmentation generator to \textit{learn} to generate the per-pixel mix weights adversarially. Different from AdvMix, the augmentation generator of AdvMix-Image outputs per-image mix weights rather than per-pixel mix weights. 
We observe that AdvMix significantly outperforms EqualMix and DirichletMix on model robustness (mPC and rPC) by a large margin, demonstrating the effectiveness of the proposed adversarial weights learning. Meanwhile, AdvMix with learnt per-pixel composition is more robust than learnt per-image composition (AdvMix-Image) since per-pixel composition could generate more diverse and fine-grained training samples to boost robustness performance, \ie, per-image mixing can hardly composite image in region-level, \eg, person and background, while per-pixel mixing with learnt weights can pay more attention to important regions. 
As illustrated in Table~\ref{tab:ablation_kd}, knowledge distillation prevents over-fitting to the induced noise patterns from creeping into the feature space and helps maintain or improve the performance on clean data.



Since we focus on how to improve robustness on \textbf{unseen} data, the corruption images in the benchmark should \emph{not} be encountered while training. However, to investigate the phenomenon if we augment the images with the test-time corruption type in the benchmark, we select one corruption type (\ie Gaussian noise, Defocus blur, Snow and Contrast) from each corruption category (\ie noise, blur, weather, and digital) as augmentation types. As illustrated in Table~\ref{tab:type}, we can observe that augmenting the training samples with test-time corruptions can boost the robustness, but the performance on clean data decreases a lot. Meanwhile, augmenting with some specific types of noises improves the performance on the target noises, but it does not always generalize to other unseen corruption types, even within the same corruption category (\eg augmentation of Snow does not contribute to improving the robustness of Fog and Brightness). By contrast, AdvMix training with stylized data achieves the best mPC. It consistently improves the performance across all corruption types, while maintaining similar clean data performance as standard training.

\subsection{Qualitative Comparison}
In Figure \ref{fig:qualitative_results}, we visualize the results of images with different types of image corruptions, \ie impulse noise, Motion blur, Brightness, and JPEG compression. 
We observe that 1) the standard pose models suffer a large performance drop on corrupted data, and 2) models trained with AdvMix perform consistently better than the baseline methods on various corruptions. 

\begin{figure}[h]
\centering
\includegraphics[width=0.45\textwidth]{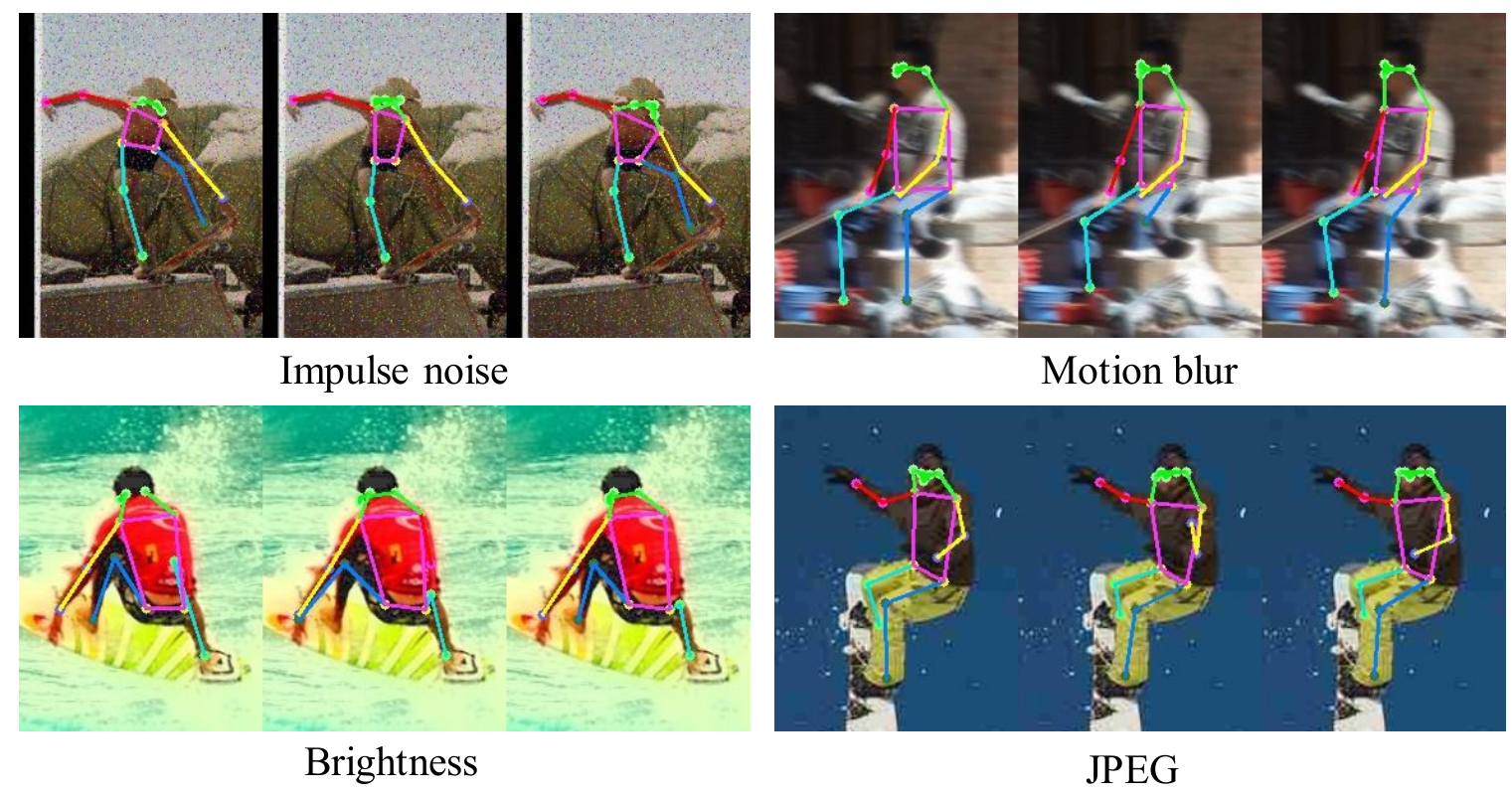}
\caption{Qualitative comparisons between HRNet without and with AdvMix. For each image triplet, the images from left to right are ground truth, predicted results of Standard HRNet-W32, and predicted results of HRNet-W32 with AdvMix.}
\label{fig:qualitative_results}
\end{figure}

\section{Conclusion}
In this paper, we propose the Robust Pose Benchmark (COCO-C, MPII-C, OCHuman-C) and rigorously evaluate the performance of current state-of-the-art models on corrupted images. Based on the benchmark, we have some interesting conclusions and examine some useful techniques to improve model robustness. We envision this work will draw the community's attention to this challenging problem and promote the development of robust pose estimators. To improve the model robustness, we propose AdvMix, a novel model-agnostic data augmentation method, to learn to mix up randomly augmented images. Our method significantly improves the robustness of most existing pose estimation models across a wide range of common corruptions while maintaining performance on clean data without extra inference computational overhead.

\noindent \textbf{Acknowledgement.} Ping Luo was supported by the Research Donation from SenseTime and the General Research Fund of HK No.27208720.

{\small
\bibliographystyle{ieee_fullname}
\bibliography{egbib}
}

\clearpage
\section*{Appendix}

\subsection*{A. Architecture of Augmentation Generator}
\begin{figure}[htb]
\centering
\includegraphics[width=0.45\textwidth]{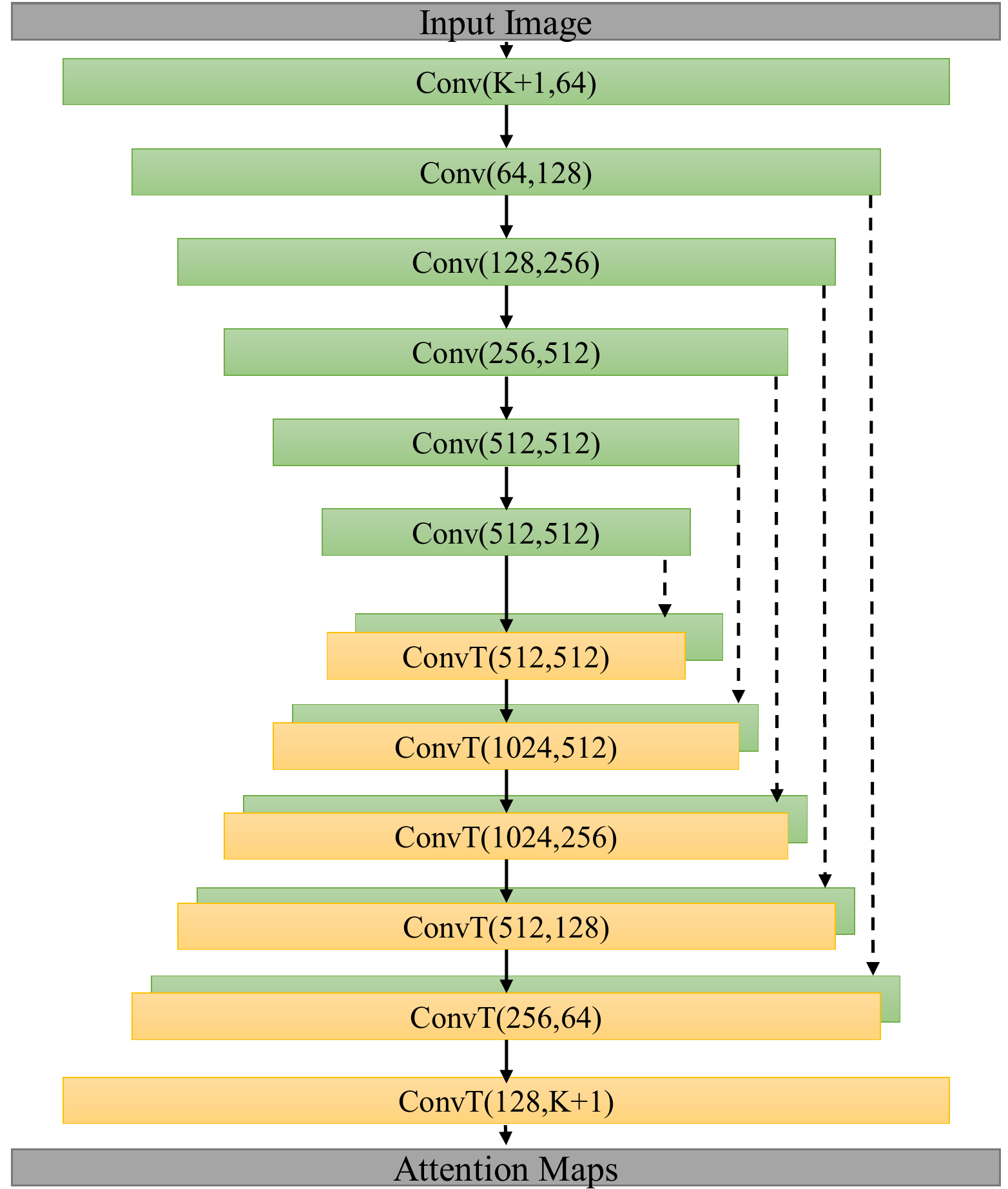}
\caption{Architecture of Augmentation Generator with input size $256 \times 192$. Conv(c,k) means the convolution block with the output channel of c, the kernel size of k. ConvT(c,k) means the transposed convolution block with the output channel of c, the kernel size of k. The kernel size and stride for all blocks are 4 and 2. The activation layer for convolution layers is ReLU, while for transposed convolution layers is LeakyReLU. The black dotted arrow lines between Conv and ConvT denote feature concatenation. }
\label{fig:auggenerator}
\end{figure}

We adopt the U-Net~\cite{Unet} architecture to build the augmentation generator, which generates the attention maps for mixing up randomly augmented images. As shown in Figure \ref{fig:auggenerator}, the augmentation generator is an encoder-decoder with skip connections in between layers, which consists of 6 convolution blocks and 6 transposed convolution blocks. To make sure the size of down-sampled features are equal to up-sampled features for concatenation, we only utilize 5 convolution blocks and 5 transposed convolution blocks for models of input size $384 \times 288$ and $128 \times 96$.

\subsection*{B. Robustness Enhancement for Different Input Sizes}
Taking HRNet-W32 as the backbone network, we conduct more experiments with different input size $128 \times 96$, $256 \times 192$, and $384 \times 288$ on COCO-C to verify the effectiveness of AdvMix for different input resolutions. As shown in Table \ref{tab:advmix_hrnet_sizes}, AdvMix improves both mPC and rPC significantly for all the three different input sizes.

\begin{table}[htb]
    \centering
    \caption{\small \textbf{Comparisons} of same backbone with different input sizes between standard training and AdvMix on COCO-C. Results are obtained with the same detection bounding boxes as \cite{hrnet}. We observe that both mPC and rPC are greatly improved, while almost maintaining performance of clean data. }
    \scalebox{0.8}{
    \begin{tabular}{c|c|c|c|c|c}
\hline
Method & Backbone &  Input size &AP$^*$ & mPC & rPC\\
\hline
Standard&HRNet-W32 & $128 \times 96$
&66.9&47.2&70.6\\
\textbf{AdvMix}&HRNet-W32& $128 \times 96$
&66.3 &48.9 &73.8 \\
\hline
Standard&HRNet-W32 & $256 \times 192$
&74.4&53.0&71.3\\
\textbf{AdvMix}&HRNet-W32& $256 \times 192$
&74.7&55.5&74.3
\\
\hline
Standard&HRNet-W32 & $384 \times 288$
& 75.7&53.7&70.9\\
\textbf{AdvMix}&HRNet-W32& $384 \times 288$
&76.2 &56.8 &74.5\\
\hline
    \end{tabular}}
    \label{tab:advmix_hrnet_sizes}
\end{table}


\subsection*{C. Visualization Results}
In Figure \ref{fig:qualitative_results}, we provide more human pose results of images with different types of image corruptions, \ie gaussian noise, motion blur, frost and contrast. For each triplet, we visualize the ground-truth (the left column), the prediction of the Standard HRNet-W32 (the middle column), and the prediction of HRNet with AdvMix (the right column). We observe that 1) the standard pose models suffer large performance drop on corrupted data, and 2) models trained with AdvMix perform consistently better than the baseline methods on various corruptions. 

\begin{figure*}[t]
\centering
\includegraphics[width=1.0\textwidth]{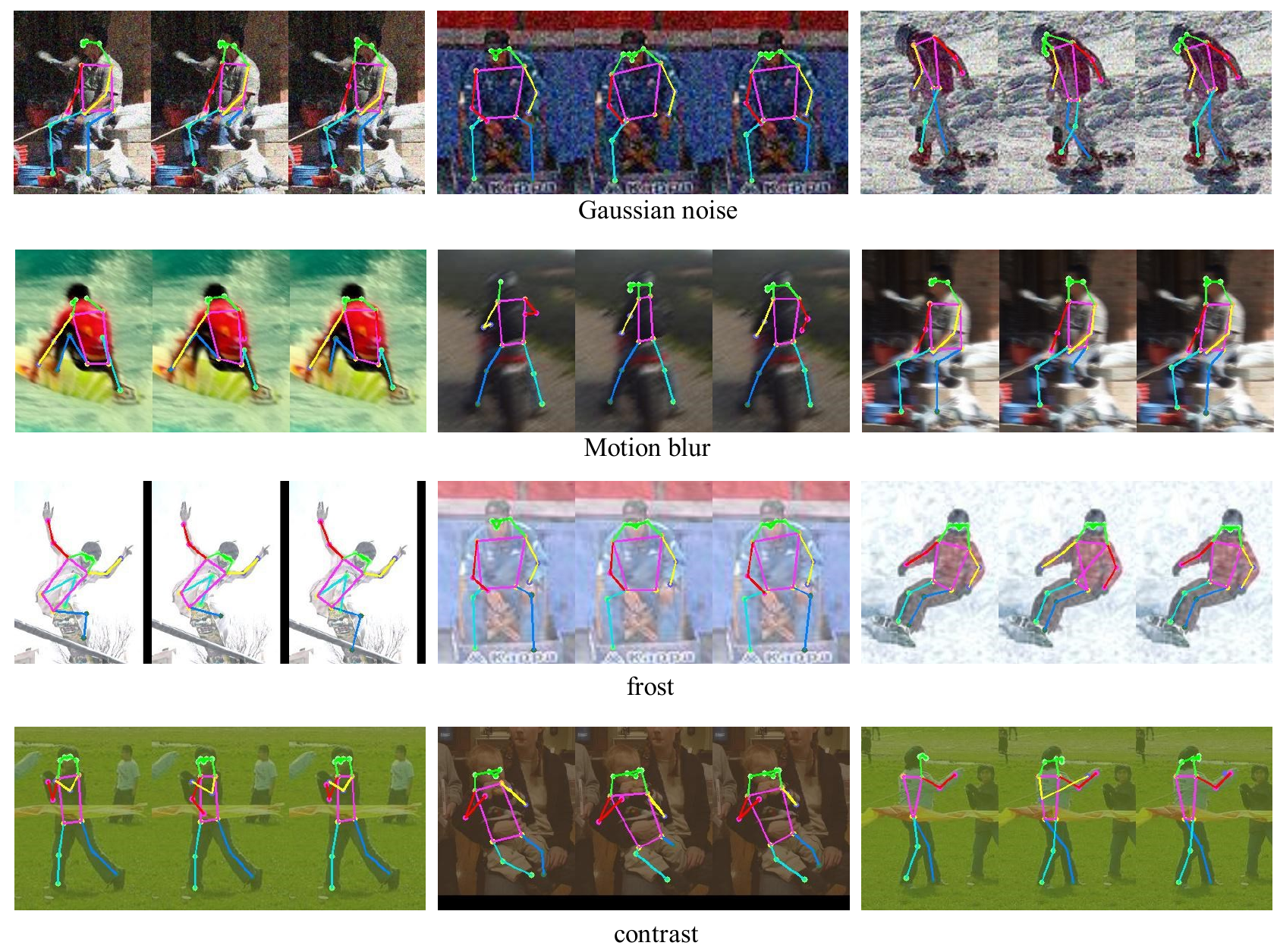}
\caption{\small Qualitative comparison between HRNet without and with AdvMix. For each image triplet, the images from left to right are ground truth, predicted results of Standard HRNet-W32, and predicted results of HRNet-W32 with AdvMix.}
\label{fig:qualitative_results}
\end{figure*}

\end{document}